\def\year{2019}\relax
\documentclass[letterpaper]{article} 
\usepackage{aaai19}  
\usepackage{times}  
\usepackage{helvet}  
\usepackage{courier}  
\usepackage{url}  
\usepackage{graphicx}  
\usepackage{booktabs}
\usepackage{tabu}
\usepackage{amssymb}
\usepackage{amsthm}
\usepackage{mathtools}
\usepackage{amsmath}
\usepackage{algorithmic}
\usepackage[ruled,vlined,linesnumbered]{algorithm2e}
\usepackage{multirow}
\usepackage{subfigure}

\def\WIDTHFIVE {0.18\textwidth}
\def\WIDTHTWO {0.48\linewidth}

\newcommand{\eg}{\emph{e.g. }}
\newcommand{\ie}{\emph{i.e. }}
\begin{document}
%
\title{Weakly Supervised Scene Parsing with Point-based Distance Metric Learning}
\author{Rui~Qian$^{1,3}$, Yunchao~Wei$^3$\thanks{Corresponding author.}, Honghui~Shi$^{2,3}$, Jiachen~Li$^3$, Jiaying~Liu$^1$ and Thomas~Huang$^3$
\\
\normalsize
$^1$Institute of Computer Science and Technology, Peking University, Beijing, P.R. China
\\
\normalsize
$^2$IBM T.J. Watson Research Center
\\
\normalsize
$^3$Image Formation and Processing Group, Beckman Institute, UIUC
\\
\tt\small
\{qianrui,liujiaying\}@pku.edu.cn, \{yunchao,hshi10,jiachenl,t-huang1\}illinois.edu
}
\maketitle
\begin{abstract}
Semantic scene parsing is suffering from the fact that pixel-level annotations are hard to be collected. To tackle this issue, we propose a Point-based Distance Metric Learning (PDML) in this paper. PDML does not require dense annotated masks and only leverages several labeled points that are much easier to obtain to guide the training process. Concretely, we leverage semantic relationship among the annotated points by encouraging the feature representations of the intra- and inter-category points to keep consistent, \ie points within the same category should have more similar feature representations compared to those from different categories. We formulate such a characteristic into a simple distance metric loss, which collaborates with the point-wise cross-entropy loss to optimize the deep neural networks. Furthermore, to fully exploit the limited annotations, distance metric learning is conducted across different training images instead of simply adopting an image-dependent manner. We conduct extensive experiments on two challenging scene parsing benchmarks of PASCAL-Context and ADE 20K to validate the effectiveness of our PDML, and competitive mIoU scores are achieved.
\end{abstract}

\section{Introduction}
\label{sec:inro}
\noindent Scene parsing is of critical importance in mid- and high-level vision. Various machine vision applications lie on the basis of detailed and accurate scene analysis and segmentation \ie outdoor driving system \cite{ess2009segmentation}, image retrieval \cite{wan2014deep} and editing \cite{tsai2016sky}. To achieve the goal of successfully recognizing objects in common scenes, datasets with careful and comprehensive labeling such as the PASCAL-Context \cite{mottaghi_cvpr14} and ADE 20K \cite{zhou2017scene} are putting forward to the community at great expense. Compared with image semantic segmentation where many semantic regions are coarsely divided into a unified background class, pixel-wise annotations for scene parsing datasets require much more efforts as more fine-grained regions (\eg wall and ground) need to be specified manually. Though current state-of-the-art methods such as PSPNet \cite{zhao2017pyramid} and Deeplab \cite{chen2018deeplab} have already obtained impressive performance on these datasets, the limitations of relying on densely-annotated data would be magnified as dozens of new circumstances \ie autonomous driving in the wild, drone navigation coming into horizon. It is both time consuming and expensive to carefully annotate a brand new dataset for each specific application. \\
\indent Attempts have been made to alleviate the annotation burden by introducing multiple weakly supervised formats \ie image-level supervision \cite{pathak2014fully}, box  supervision \cite{dai2015boxsup} and scribble supervision \cite{lin2016scribblesup}. In the mean time, as suggested in \cite{bearman2016s}, to annotate a single pixel for each instance is a natural scheme for human reference and the cost could be greatly alleviated. Furthermore, Bearman's work tackles the easier semantic segmentation and focuses more on analyzing the effectiveness of the point-based scheme itself by analyzing the annotation quality and taking the interaction with annotators as an important factor in the loss function. We believe that the potential of point-based supervision has not been well explored and also, the more challenging task of point-based scene parsing has remained untouched. \\
\indent Thus, this paper serves as an initial attempt to explore the possibility of point-guided weakly supervised scene parsing. Being given only one semantic annotated point per instance, we propose a novel point-based distance metric learning method (PDML) to tackle this challenging task. PDML leverages semantic relationship among the annotated points by encouraging the feature representations of the intra- and inter-category points to keep consistent. Points within the same category are optimized to share more similar feature representations and oppositely, features of points from different categories are optimized to be more distinct. We implement this optimizing procedure by utilizing a distance metric loss, which collaborates with the point-wise cross-entropy loss to optimize the whole deep neural network. More important, different from current weakly supervised methods whose solutions are constrained in a single image, we conduct distance metric learning across different training images, so that the limited human annotated points can be fully exploited. Extensive experiments are performed on two challenging scene parsing benchmarks: PASCAL-Context and ADE 20K. We achieve the mIoU score of 30.0\% on PASCAL-Context, which is impressive compared to the result of 39.6\% from fully supervised scheme by using only $7.2\times 10^{-5}$ the number of annotated pixels. And we achieve mIoU of 19.6\% on ADE 20K, while the SegNet \cite{badrinarayanan2017segnet} produces roughly 21\%  under full annotation of this dataset.

In conclusion, the major contributions of this paper lie in the following aspects:
\begin{itemize}
\item We are the first to deal with the task of point-based weakly supervised scene parsing.
\item We propose a novel deep metric learning method PDML which optimizes the intra- and inter-category embedding feature consistency among the annotated points.
\item PDML is performed across different training images to fully exploit the limited annotations, which is very novel compared to traditional intra-image methods.
\item Our method has competitive performance both qualitatively and quantitatively on PASCAL-Context and ADE 20K scene parsing dataset.
\end{itemize}

\section{Related Work}
\label{sec:rw}
\subsection{Weakly Supervised Semantic Segmentation}
\paragraph{Image-level Annotations} Image-level annotations by just naming the objects and staff are easy and natural to obtain. \cite{pathak2014fully} explore multiple instance learning for semantic segmentation. \cite{papandreou2015weakly} propose dynamic prediction of foreground and background by using Expectation-Maximization algorithm. \cite{kolesnikov2016seed} introduce the new loss function of seed, expand and constrain. \cite{wei2017stc} utilize a simple to complex framework to further improve the performance. And object region mining and localization is experimented in \cite{wei2016learning},  \cite{wei2017object}, \cite{zhang2018adversarial} and \cite{hou2018self}. And \cite{wei2018revisiting} discuss the effect of dilated convolution in this task. However, the methods listed below all focus on object-based semantic segmentation while no attempts have been made to deal with the more difficult scene parsing. Recently \cite{zhang2018hierarchical} tackle image-description guided scene parsing but investigating on parsing a scene image into a structured configuration.
\vspace{-3.5mm}
\paragraph{Region-specified Annotations}
Compared to image-level annotations where no explicit location related information are given, multiple attempts have been made to provide various region-specified semantic supervision. Annotated bounding boxes are utilized in \cite{dai2015boxsup} and \cite{papandreou2015weakly}. \cite{lin2016scribblesup} use scribbles as the supervision information and graphic models are used in optimization. Furthermore, \cite{bearman2016s} have a similar setup with us but target at semantic segmentation. Also, they focus more on analyzing point supervision regime itself by comparing the annotation time, error rate as well as quality with other supervision regimes. Their method takes the confidence of annotators as a parameter in the loss function and transfers image-level annotations to objectness priors by utilizing another model pretrained on non-overlapping datasets. In comparison, we put attention on the more difficult task of scene parsing. We do not use any additional data and focus on exploring the cross-image semantic relations to boost the scene parsing performance. 

\subsection{Deep Metric Learning}
Deep metric learning on embedding features has been explored in various tasks such as image query-and-retrieval \cite{oh2016deep}, face recognition \cite{schroff2015facenet} and verification \cite{ming2017simple}. It has also been applied in semantic instance segmentation \cite{fathi2017semantic} and grouping \cite{kong2018recurrent}. More recently, \cite{liu2018deep} use metric learning as a fast and efficient way for training a semantic segmentation network. 

\begin{figure*}[hbtp]
  \begin{center}
    \includegraphics[width=\textwidth]{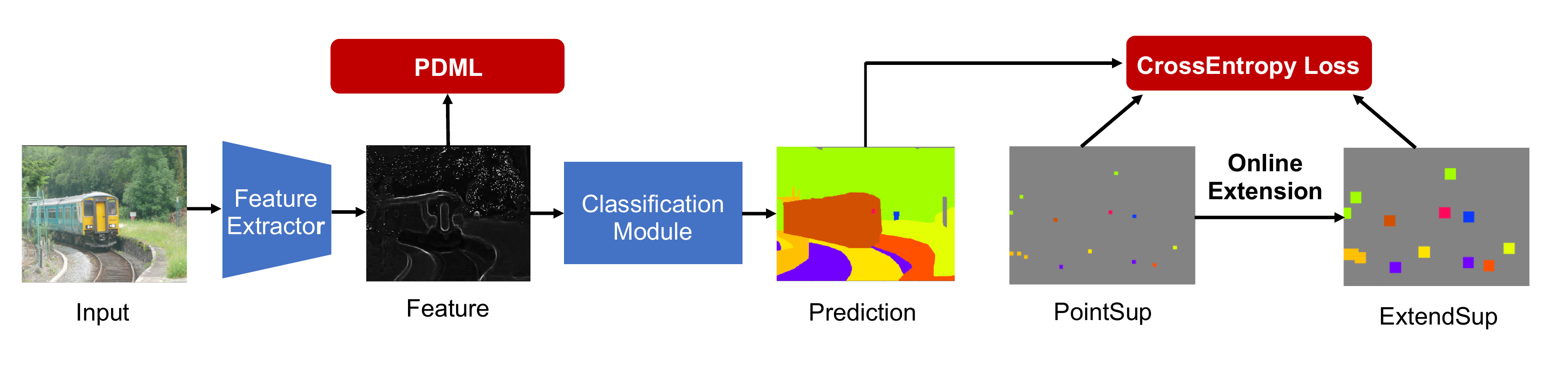}
  \end{center}
  \caption{The pipeline of our proposed algorithm.}
  \label{fig:pipe}
  \vspace{-3mm}
\end{figure*}

\section{Proposed Method}
\label{sec:method}

\subsection{Motivation}
Compared to full, box and scribble supervision regime, point-based supervision data is most natural for human reference and easiest to obtain. However, as illustrated in Table \ref{tab:num}, the annotation number of pixels in single image is too tiny to train a neural network efficiently.
\vspace{-0.3cm}
\begin{center}
\begin{table}[!htbp]
\begin{tabular}{c|ccc}
\toprule[2pt]
Annotation Method & Full & ScribbleSup & PointSup\\
\midrule[1pt]
Anno. pixel/image&170k&1817.48&\textbf{12.26}\\
\bottomrule[2pt]
\end{tabular}
\caption{Comparison of the average annotated pixel number per image of different methods on PASCAL-Context training dataset.}
\label{tab:num}
\end{table}
\end{center}
\vspace{-0.8cm}

\indent Consider the limitation of supervision information and inspired by recent deep metric learning methods \eg \cite{liu2018deep}, we focus on exploring the relationship between feature representations of annotated pixels. While all current methods try to optimize embedding feature distances within a single image, we apply a novel method by forming triples from the embedding vectors of annotated pixels across images and optimizing the feature consistency within the triples by distance metric learning. In each triple, two embedding vectors belong to the same category and we name them as a positive pair. The other is from a different category and it forms a negative pair with one element in the positive pair. We minimize the distances between positive pairs and maximize those between negative pairs on inter-image level. There are at least two reasons for doing so:

\begin{itemize}
\item Objects and stuff which have similar feature representations before the classification module would be more likely to be specified into the same class. Oppositely, embedding features from different categories, if being distinct enough with each other, are more easier for the classifier to distinguish.
\item Under the point-based regime, most annotated pixels in one image come from different categories. Simply optimizing distances between negative pairs would not help training. While extending to inter-image level, balanced number of positive and negative pairs can be obtained.
\end{itemize}

Furthermore, different from image-level weakly supervised methods relying heavily on the saliency maps generated by pretrained specific models, scribble and box guided methods depending on being optimized iteratively, our method does not require any additional data and can be learned in an end-to-end manner.

\subsection{Overview}

Figure \ref{fig:pipe} shows the pipeline of our algorithm. Similar to Deeplab \cite{chen2018deeplab}, we utilize ResNet101 \cite{he2016deep} pretrained on ImageNet as the backbone of our feature extractor. Atrous convolutions are adopted to increase the receptive field and reduce the degree of signal downsampling. Given a batch of input images, we first use the feature extractor to form deep embedding features. The features are then assigned to two streams. The first stream is feeding into the point-based distance metric learning module where the embedding feature vectors across different images are optimized towards learning representation consistency. The second stream is feeding into a fully-convolutional classification module to generate the pixel-wise prediction. At the mean time, online extension is performed to dynamically gather pixels with high classification confidence and tight spatial relationship to the original annotated pixels to form the extended label. The point-wise cross-entropy loss is calculated between the classification results and the two labels. 

\begin{figure*}[h]
  \begin{center}
    \includegraphics[width=\textwidth]{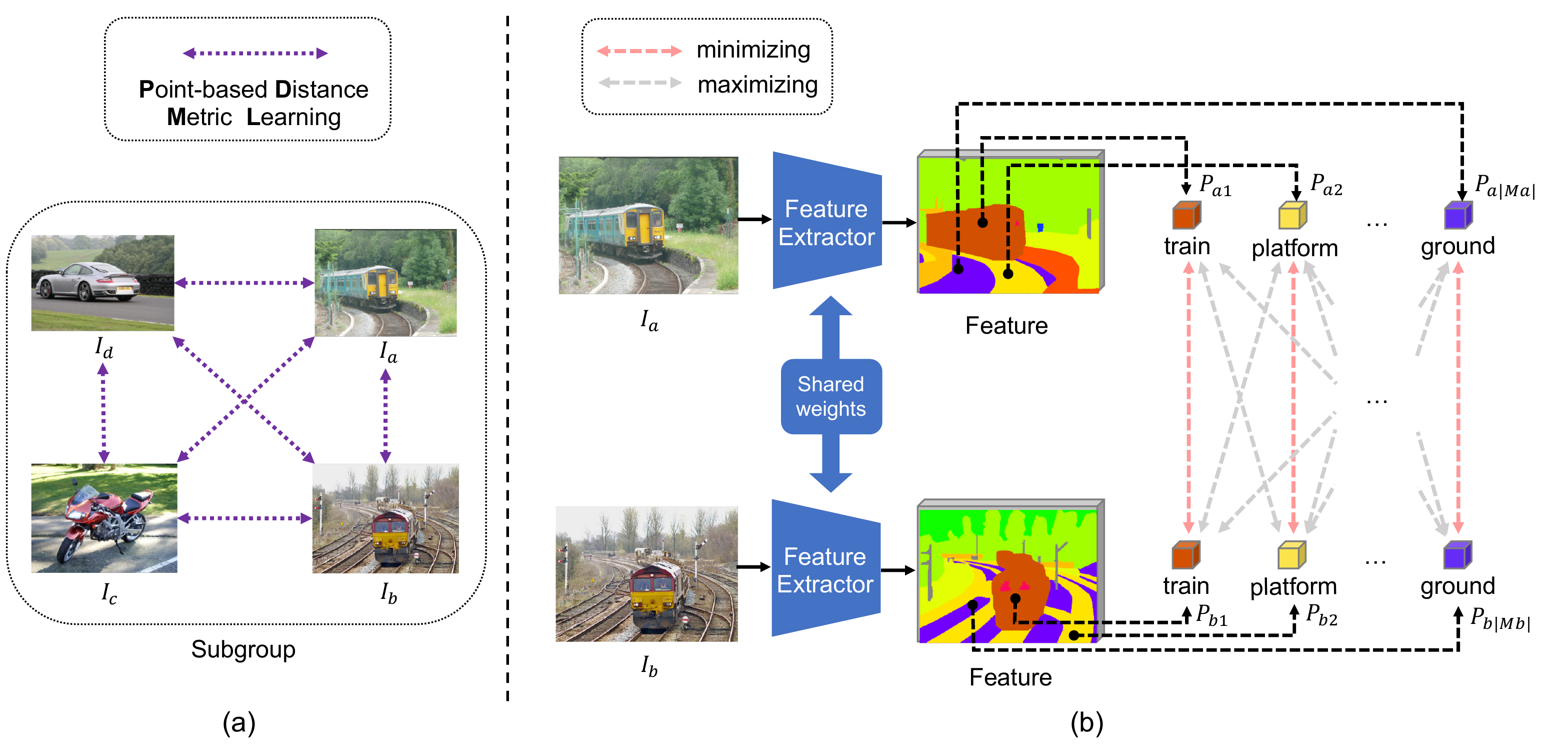}
  \end{center}
  \caption{The architecture of our dense PDML. As shown in part (a), we first densely form image pairs in small subgroups divided from the training dataset. Then for each image pairs, as shown in part (b), we perform densely distance optimizing between the embedding vectors of annotated pixels. Best viewed in color.}
  \label{fig:network}
\end{figure*}

\subsection{Point Supervision}
Point supervision (PointSup) serves as the baseline of our method. Let $\mathcal{S}_p$ donate the set of training images with point-wise annotations. Then we have $\mathcal{S}_p = \bigcup_{w=1}^{N}\{(I_{w}, M_{w})\}$ where $I_w$ is the $w$th image in the training set. $M_w$ is the corresponding pseudo annotated mask, where only several pixels are annotated with semantic labels. Let $g(f(I_w; \theta_{f}); \theta_{g})$ donate our segmentation module where $(f, \theta_{f})$, $(g, \theta_{g})$ refer to the feature extraction module and its parameters, classification module and its parameters, respectively. The objective function is

\begin{equation}
\min\limits_{\theta_f, \theta_g}\sum_{w=1}^{N}J_w(g(f(I_w; \theta_{f}); \theta_{g})),
\end{equation}
and then consider the class label set of the current image as $\mathcal{C}$, let the the conditional probability generated by the classification module of any label ${c}\in{\mathcal{C}}$ at the any location $u$ as $g_{u,c}(f(I_w; \theta_{f}); \theta_{g})$, then we get

\begin{equation}
J_w(g(f(I_w; \theta_{f}); \theta_{g})) \! = \! -
\frac{\sum\limits_{{c}\in{\mathcal{C}}}\!\sum\limits_{{u}\in{M^{c}_{w}}}\!\log{g_{u,c}(f(I_w; \theta_{f}); \theta_{g})}}{{\sum\limits_{{c}\in{\mathcal{C}}}|M^{c}_{w}|}},
\end{equation}
where $|*|$ refers to number of annotated pixels in $*$.

\subsection{Distance Metric Learning}
To optimize feature representations of annotated pixels to keep consist, we aim at minimizing distances between positive pairs and maximizing those between negative pairs. However, it is hard to find positive pairs within a single image. Only optimizing negative pairs would do no help but make the loss hard to converge. By extending to inter-image level, we could obtain balanced number of two kinds of pairs as illustrated in Figure \ref{fig:network}. Let $s$ be a subgroup of the training set $\mathcal{S}_p$, then $s = \{(I_a, M_a), (I_b, M_b), ...\}$. For each image \eg $I_a$ in the subgroup, we could define the embedding vector set $E_{a} = \bigcup_{i=1}^{|M_a|}\{P_{ai}\}$, where $P_{ai}$ is the feature vector corresponding to the $i$th annotated pixel in image $a$. Suppose for three different feature vectors $P_{ai}$, $P_{bj}$, $P_{bk}$, where $P_{ai}$ shares the same category with $P_{bj}$ and different from $P_{bk}$, we apply the loss $L_t$ as
\begin{equation}
\begin{split}
L_t(P_{ai}, P_{bj}, P_{bk}) = &\alpha L_p(P_{ai}, P_{bj}) + \beta L_n(P_{ai}, P_{bj}, P_{bk}),
\end{split}
\end{equation}
where $L_p(P_{ai}, P_{bj})$, corresponding to the red dotted line in Figure 2(b) (\ie ($P_{a1}, P_{b1}$)), can be expressed as
\begin{equation}
\begin{split}
\|P_{ai}-P_{bj}\|_{2},
\end{split}
\end{equation}
which aims at minizing the L2-norm distance between same-category embedding vectors. And $L_n(P_{ai}, P_{bj}, P_{bk})$, corresponding to the combination of one gray and one red dotted line in Figure 2(b) (\ie ($P_{a1}, P_{b1}$) and ($P_{a1}, P_{b2}$)), can be expressed as
\begin{equation}
\begin{split}
\max(\|P_{ai}-P_{bj}\|_{2} - \|P_{ai}-P_{bk}\|_{2} + m, 0), 
\end{split}
\end{equation}
which aims at maximizing the gap between $\|P_{ai}-P_{bj}\|_{2}$ and $\|P_{ai}-P_{bk}\|_{2}$. $m$ is a constant value and only when the gap is within this value would the the triple embedding vectors be optimized. Two hyper-parameters $\alpha$ and  $\beta$ are used to balance the the effect of $L_p$ and $L_n$. We set $m = 20$, $\alpha = 0.8$, $\beta = 1$ in practice. And the algorithm of optimizing dense PDML is expressed in Algorithm \ref{alg1}.

\begin{algorithm}
\DontPrintSemicolon
\SetAlgoLined
\While{$|S_p|\geq 0$}
{
    {Extract supgroup $s$ from $S_p$;}\\
    \For{$(I_n, M_n) \in s$}
      {
      \For{$(I_m, M_m) \in s - (I_n, M_n)$}
      {
      	$Generate \ E_n, E_m$ \\
      	\For{$P_{ni} \in E_n$}
     	 {
            $E_{pos}$ = \{\}, $E_{neg}$ = \{\}\\
            \For{$P_{mj} \in E_m$}
            {
              \eIf{$class(P_{mj}) = class(P_{ni})$}
              {
                  $E_{pos} = E_{pos}\cup \{P_{mj}\}$
              }
              {
                  $E_{neg} = E_{neg}\cup \{p_{mj}\}$
              }
           	}
            \While{$|E_{pos}| \geq 0 \ and \ |E_{neg}| \geq 0$}
            {
            	{$select \ P_{pos} \in E_{pos}, P_{neg} \in E_{neg}$}\\
             	{$\min L_t(P_{ni}, P_{pos}, P_{neg}) $}\\
                {$E_{pos} = E_{pos} - \{P_{pos}\}$}\\
                {$E_{neg} = E_{neg} - \{P_{neg}\}$}\\
           	}
          }
      }
    }
    $S_P = S_P - s$
}
\caption{Optimizing Procedure for PDML}
\label{alg1}
\end{algorithm}

\subsection{Online Extension}
To further improve the performance, we put attention on gathering more pixels during the training process. Previous works such as superpixel \cite{achanta2012slic} and K-Means clustering \cite{ray1999determination} have explored the possibility of gathering pixels by measuring the similarity of low-level features. However, both methods have obvious drawbacks under the point-based scene parsing regime by gathering lots of wrong pixels. And from experiments we find that wrong pixels would greatly degrade the performance of the network. More detailed analyses would be found in the section of discussion. To tackle this issue, we adopt a simple but accurate online extension method to collect more pixels with low false positive rate to extend the annotation data. Take $(I_a, M_a) \in S_p$ as an example, we now assume the current weights of feature extractor $f$ and classification module $g$ as $\theta_{f1}$ and $\theta_{g1}$, then we could specify the new label candidate $M_{a\_score}$ by judging the pixel-wise classification score:
\begin{equation}
M_{a\_score}\!=\!\bigcup_{u} \{\max(g_{u,c\in C}(f(I_a; \theta_{f1}); \theta_{g1}))\!>\!thr\},
\end{equation}
where $thr$ is a threshold to filter the pixels. In other words, for every location $u$ in the input image, if the max classification score of this pixel is greater than the threshold and the corresponding class is within the class label set of this image, it would be chosen for the extended label. From another perspective, we believe that pixels with close spatial distance to the annotated pixels are more likely from the same category. Thus we extend each annotated pixel in $M_a$ to a $5\times5$ square to form the second  label candidate $M_{a\_region}$. Then we generate the final extended label $M_{a\_extend}$ by just selecting the candidates appearing in both schemes as:
\begin{equation}
M_{a\_extend} = M_{a\_score} \cap M_{a\_region}.
\end{equation}

\section{Experimental Results}
\label{sec:results}
\subsection{Implementation Details}
\paragraph{Dataset and Evaluation Metrics} Our proposed model is trained and evaluated on two challenging scene parsing datasets: PASCAL-Context \cite{mottaghi_cvpr14} and ADE 20K \cite{zhou2017scene}, as shown in Table \ref{tab:1}. In PASCAL-Context, the most frequent 59 classes are used and others are divided into a unified background class. In ADE 20K, we adopt the annotation of 150 meaningful classes. We generate one pixel annotation for each instance in each image. The performances are quantitatively evaluated by pixel-wise accuracy and mean region intersection over union (mIoU). 

\begin{center}
\begin{table}[htbp]
\begin{tabular}{cccc}
\toprule[2pt]
Dataset&\# Training &\# Eval & Pixel/Image\\
\midrule[1pt]
PASCAL-Context&4998&5105&12.26\\
ADE20K&20210&2000&13.96\\
\bottomrule[2pt]
\end{tabular}
\caption{Statistical results of two datasets. \textbf{Pixel/Image} illustrates the average number of pixels annotated in one image in the training dataset.}
\label{tab:1}
\end{table}
\end{center}
\vspace{-12mm}
\paragraph{Training Setting}
We utilize ResNet101 \cite{he2016deep} with the modification of atrous convolution as the backbone of our feature extractor. Weights pretrained on ImageNet are adopted to initialize. During training, we take a mini-batch of 16 images and randomly crop patches of the size of $321\times321$ from original images. We use the optimizer of SGD where momentum is set to 0.9 and the weight decay is 0.0005. The initial base learning rate is set to 0.00025 for parameters in the feature extraction layers and ten times for parameters in the classification module. Both learning rate will be decayed under the scheme of $base\_lr*(1 - \frac{epoch}{max\_epoch})^{0.8}$. All the experiments are conducted on two NVIDIA V100 GPUs. Our code will be available publicly.

\vspace{-1.5mm}
\subsection{Quantitative and Qualitative Results}
We evaluate different methods quantitatively by using pixel accuracy and mIoU which describes the the precision of prediction and the average performance among all classes, respectively. We run multiple experiments to determine the effects of the three parts of our proposed method: PointSup, PDML and online extension. To make a more comprehensive understanding of the effect of our method, we also train our network with fully-annotated label. The quantitative results are shown in Tables \ref{tab_re:1} and \ref{tab:class} (note we omit \% of mIoU for simplicity). On PASCAL-Context, the full supervision setting could yield a 39.6\% mIoU and our method, with only $7.2\times 10^{-5}$ the number of annotated pixels, could obtain the 30.0\% mIoU performance. On ADE 20K, we achieve the mIoU of 19.6\%, while the SegNet \cite{badrinarayanan2017segnet} achieves roughly 21\% mIoU under full supervision. And the qualitative results are shown in Figures \ref{fig:compare1} and \ref{fig:compare2}. Our final method combining point supervision, distance metric learning and online extension has the best scene parsing quality subjectively.

\begin{table}[htbp]
\centering
\resizebox{\linewidth}{28mm}{
\begin{tabular}{cccc|cc}
\toprule[2pt]
\multicolumn{4}{c}{Method} & \multicolumn{2}{c}{Metrics} \\
\midrule[1.5pt]
FullSup& PointSup& PDML& Online Ext.&  mIOU& Pixel Acc\\
\midrule[1.5pt]
\multicolumn{6}{c}{PASCAL-Context validation dataset} \\
\midrule[1.5pt]
${\surd}$& & & & 39.6& 78.6\%\\
\midrule[1pt]
 & ${\surd}$& & & 27.9& 55.3\%\\
 & ${\surd}$& ${\surd}$& & 29.7&57.5\% \\
 & ${\surd}$& ${\surd}$& ${\surd}$& \textbf{30.0}& \textbf{57.6\%}\\ 
 \midrule[1.5pt]
\multicolumn{6}{c}{ADE 20K validation dataset} \\
 \midrule[1.5pt]
${\surd}$& & & & 33.9& 75.8\%\\
\midrule[1pt]
 & ${\surd}$& & & 17.7& 58.0\%\\
 & ${\surd}$& ${\surd}$& & 19.0& 59.0\%\\
 & ${\surd}$& ${\surd}$& ${\surd}$& \textbf{19.6}& \textbf{61.0\%}\\
 
\bottomrule[2pt]
\end{tabular}}
\caption{Quantitative results on PASCAL-Context and ADE 20K validation dataset.}
\label{tab_re:1}
\end{table}

\section{Discussion}
\label{sec:diss}
\subsection{Analysis of PDML}
\begin{figure}[hbtp]
\centering
\includegraphics[width=0.8\linewidth]{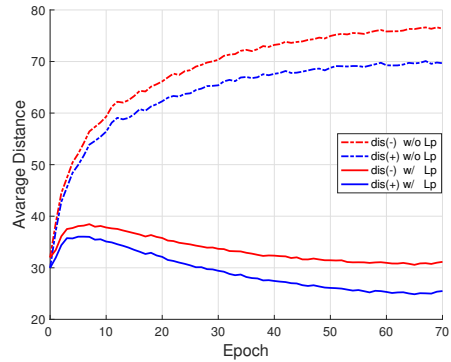}
\caption{Comparison of the average value of $dis(+)$ and $dis(-)$ on whether using $L_p$ constrain in the loss function.}
  \label{fig:lp}
\end{figure}

\paragraph{Loss function} Recall that we apply the loss $L_t = \alpha L_p + \beta L_n$ to optimize the consistency of embedding vectors. We name the distance between the positive pairs as $dis(+)$ and that between negative pair as $dis(-)$. $L_p$ aims at constraining $dis(+)$ and $L_n$ aims at increasing the gap between $dis(+)$ and $dis(-)$. We argue that $L_p$ is very import in the whole scheme, as shown in Figure \ref{fig:lp}. With only applying $L_n$, though the gap between $dis(+)$ and $dis(-)$ is optimized to be larger, the absolute values increase greatly at the mean time which leads to great performance drop. And by applying the constrain of $L_p$, the absolute values of distances remain at the normal scale and the gap is also optimized to make it easier for the classification module to distinguish.

Also, we visualize the distribution of the pair numbers of each distance value in training procedure to demonstrate the effectiveness of $L_t$. At the first epoch, the distributions of distances of positive and negative pairs are almost symmetrical. During the training process, an obvious peak shift can be observed. The peak of the distances between positive pairs moves towards the origin of the coordinate axis and peak of the distances between negative pairs moves oppositely.

\begin{figure}[htbp]
\centering
\subfigure[Epoch = 1]{
\includegraphics[width=\WIDTHTWO]{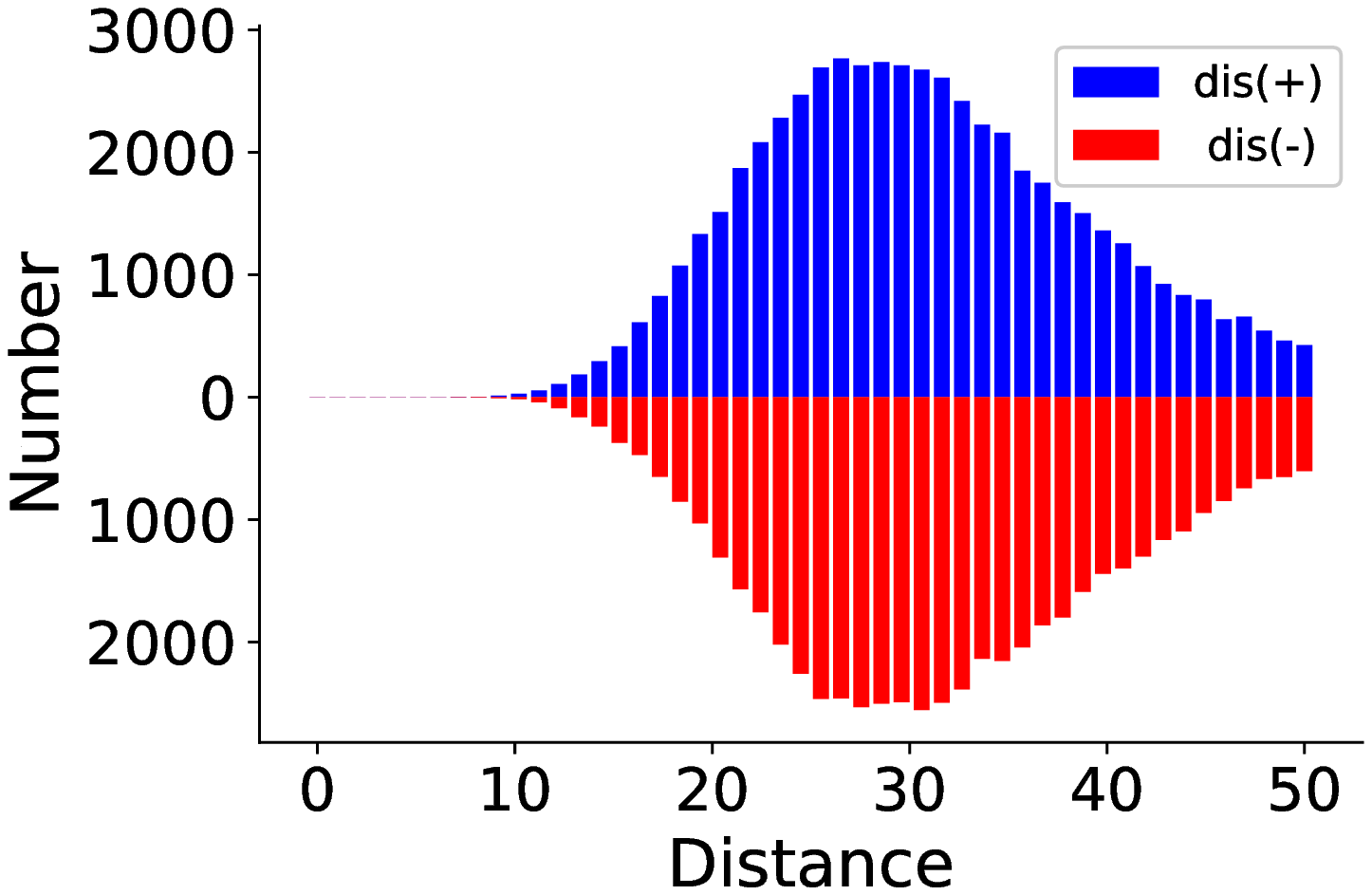}}
\subfigure[Epoch = 20]{
\includegraphics[width=\WIDTHTWO]{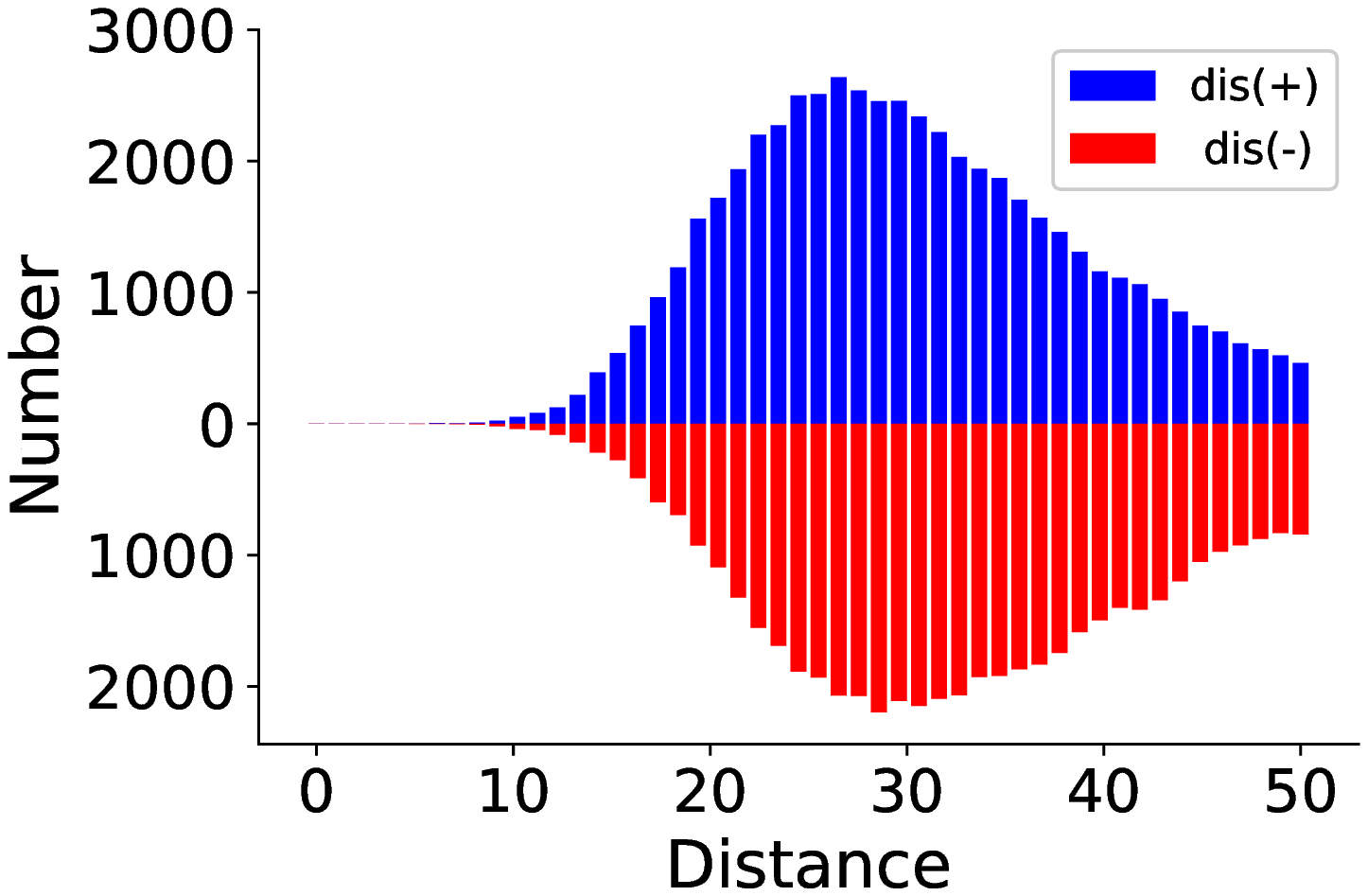}}
\subfigure[Epoch = 40]{
\includegraphics[width=\WIDTHTWO]{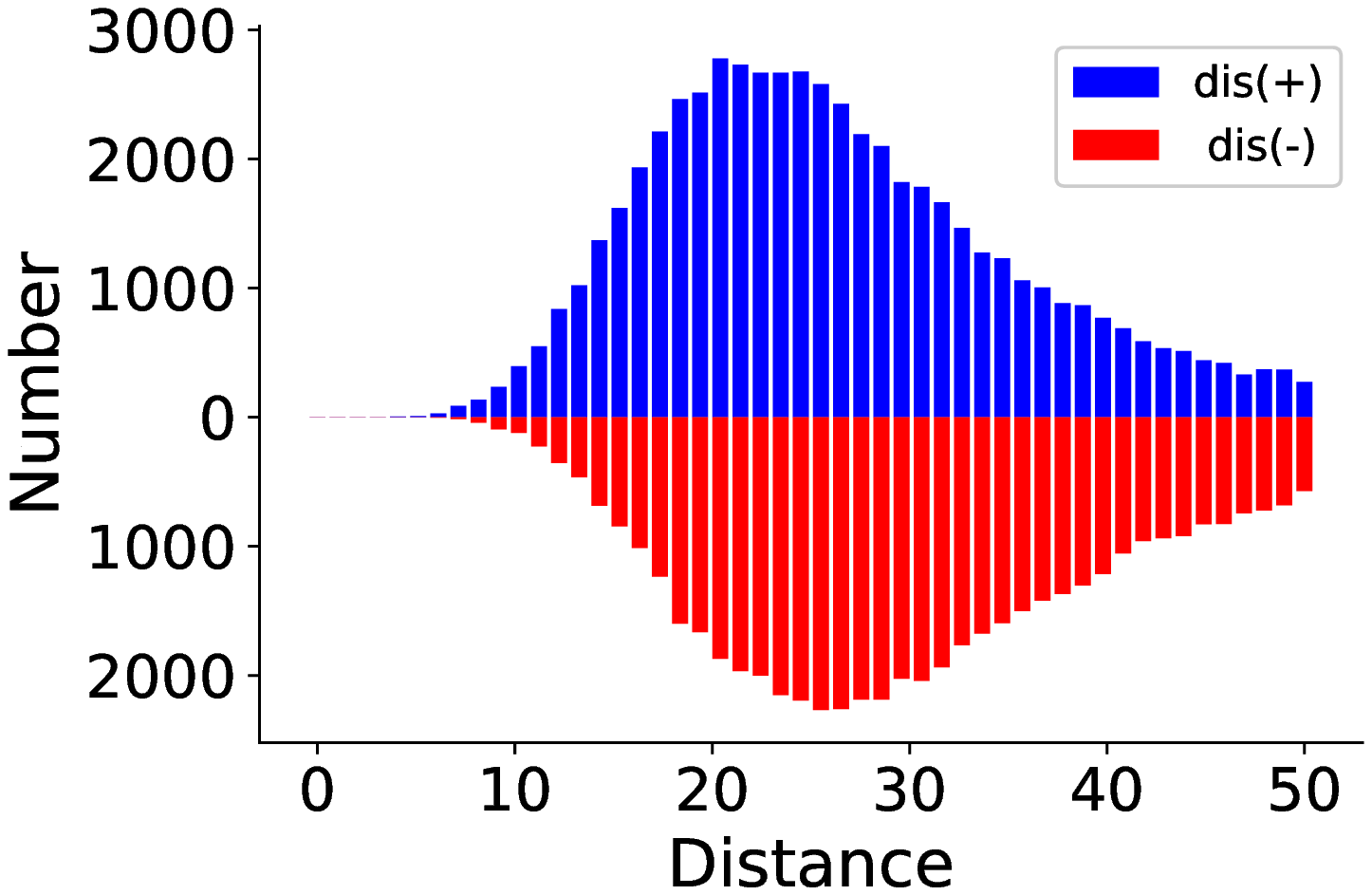}}
\subfigure[Epoch = 60]{
\includegraphics[width=\WIDTHTWO]{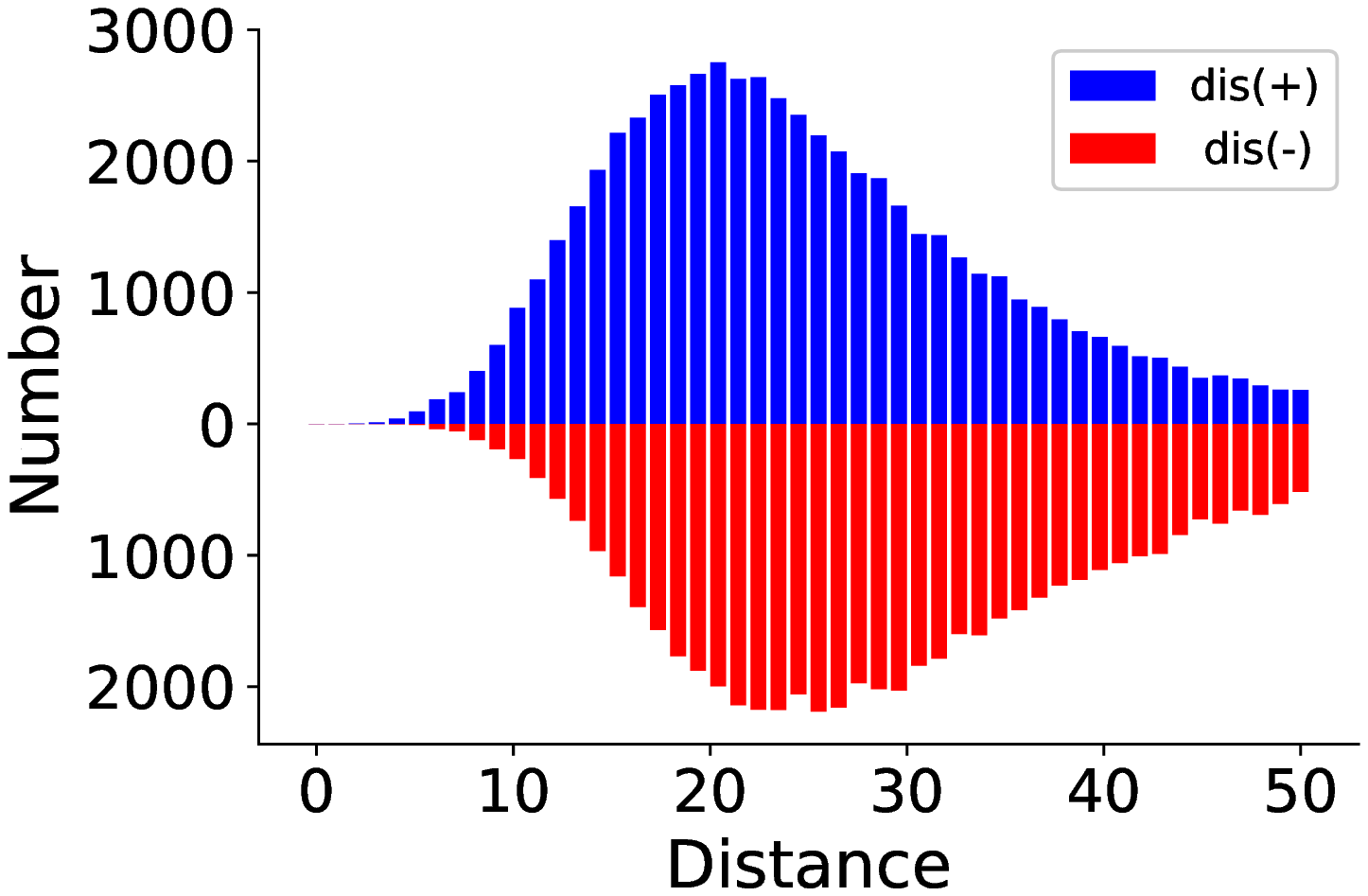}}

\caption{Summary of the distribution of the pair numbers of each distance value. An obvious peak shift can be observed.}
\label{fig:lp2}
\end{figure}
\vspace{-0.5cm}
\paragraph{Hyper-parameters} We have three hyper-parameters in the loss function: $m$, $\alpha$, $\beta$. Margin $m$ is set for a mining purpose. A small margin would limit the number of optimizing embedding vector triples and make it less effective for classification. And a large margin size, bringing too many triples for optimization, would cause the training loss being hard to converge. A moderate margin value of 20 would produce the best performance. Loss weights $\alpha$ and $\beta$ are used to balance the effect of $L_p$ and $L_n$. We set $\beta$ to 1 and adjust $\alpha$ correspondingly. A small $\alpha$, similar to Figure \ref{fig:lp}, can not constrain the absolute value of $dis(+)$ and $dis(-)$ to remain at the normal scale and would lead to poor performances. While a large $\alpha$ would weaken the effect of maximizing the distance between $dis(+)$ and $dis(-)$ and make it hard to distinguish embedding vectors from different categories. A moderate value of 0.8 achieves the best performance.

\begin{table}[hbtp]
\centering
\resizebox{0.60\linewidth}{23mm}{
\begin{tabular}{ccc|cc}
\toprule[2pt]
\multicolumn{3}{c}{Hyper-parameters} & \multicolumn{2}{c}{Metrics} \\
\midrule[1pt]
$\alpha$ & $\beta$ & m & mIoU & Pixel Acc \\
\midrule[1pt]
0.8 &1 &8 &29.2 &57.2\%  \\
0.8 &1 &12 &29.5 &57.0\% \\
0.8 &1 &16 &29.4 &57.1\% \\
0.8 &1 &\textbf{20} &\textbf{29.7} &\textbf{57.5\%} \\
0.8 &1 &24 &29.4 &56.9\% \\
\midrule[1pt]
0.0 &1 &20 & 26.2& 53.5\%\\
0.4 &1 &20 & 28.6& 55.8\%\\
\textbf{0.8} &1 &20 &\textbf{29.7} &\textbf{57.5\%} \\   
1.2 &1 &20 & 29.5& 57.1\%\\   
1.6 &1 &20 & 29.4& 57.0\%\\
\bottomrule[2pt]
\end{tabular}}
\caption{Comparisons of the performance of different hyper-parameter settings on PASCAL-Context validation dataset.}
\label{tab:margin}
\end{table}
\subsection{Analysis of Extension Method}
We compare our online label extension method with other frequently used clustering method.

\noindent\textbf{Superpixel} Superpixel is used to cluster pixels by taking low-level feature similarity into consideration. We set the number of superpixels in the range of 50 to 200 per image depending on the number of annotated pixels. Each annotated pixel would be extended to the corresponding superpixel covering it. 

\noindent\textbf{K-Means} We perform K-Means clustering on the feature representations of the input images. The annotated pixels are set to be the initial clustering centers and we set the maximum iteration time to be 300. 

\noindent\textbf{Score and Region} Recall in the online extension part, we use two methods to generate new label candidates. The first is using a score thresholding and we name this extension method as \textbf{score}. And another is simply extending every pixel in the current label to a $5\times5$ square with the original one as the center. We name this method as \textbf{region}. \\
\indent On the basis of the weight obtained by PDML, we implement different extension methods and their results are shown in Table \ref{tab:extend}. The pixel-wise extension accuracy is of critical importance in influencing the performance. Superpixel and region have better extension accuracies and have better performances correspondingly. We also test various thresholds for the score scheme. Our method taking both score with the threshold of 0.7 and region into consideration has an accuracy of 98.2\% and the best testing performance. 
\vspace{-0.2cm}
\begin{table}[htbp]
\centering
\resizebox{\linewidth}{19mm}{
\begin{tabular}{cccc|c|cc}
\toprule[2pt]
\multicolumn{5}{c}{Extension Method}& \multicolumn{2}{c}{Metrics} \\
\midrule[1.5pt]
Superpixel& K-Means& Score&  Region& Extension Acc& mIoU& Pixel Acc\\
\midrule[1pt]
${\surd}$& & & & 83.6\%& 26.7& 52.9\%\\
 & ${\surd}$& & &12.1\%& 12.9& 36.3\%\\
 & & ${\surd}(0.7)$& & 56.9\%& 16.7& 40.3\% \\
 & & & ${\surd}$& 97.6\%& 29.7& 57.5\%\\ 
\midrule[1pt]
 & & ${\surd}(0.5)$& ${\surd}$& 97.7\%& 29.7& 57.5\% \\
 & & ${\surd}(0.6)$& ${\surd}$& 98.0\%& 29.8& 57.4\% \\
  & & ${\surd}(0.7)$& ${\surd}$& \textbf{98.2\%} &\textbf{30.0}& \textbf{57.6\%}\\
 & & ${\surd}(0.8)$& ${\surd}$& 98.3\%& 29.4& 57.1\% \\
 & & ${\surd}(0.9)$& ${\surd}$& 98.4\%& 29.4& 56.9\% \\   
\bottomrule[2pt]
\end{tabular}}
\caption{Comparisons of different extension method on PASCAL-Context validation dataset.}
\label{tab:extend}
\vspace{-1mm}
\end{table}

\begin{table*}
\centering
\footnotesize
\resizebox{\textwidth}{33mm}{
\begin{tabular}{ccccccccccccc}
\toprule[2pt]
\textbf{PointSup}&\textbf{PDML}&\textbf{Onlin Ext.} & plane & bicycle & bird & boat & bottle & bus & car & cat & chair& cow\\
\midrule[1pt]
${\surd}$&&&41.8&39.8&52.7&32.6&39.3&50.3&48.9&64.5&17.3&\textbf{44.0}\\
${\surd}$&${\surd}$&&\textbf{47.3}&\textbf{46.9}&55.1&34.0&46.6&50.7&\textbf{51.4}&67.1&\textbf{21.9}&40.1\\
${\surd}$&${\surd}$&${\surd}$&45.7&44.5&\textbf{56.7}&\textbf{34.2}&\textbf{46.8}&\textbf{50.9}&50.2&\textbf{67.1}&21.3&41.8 \\ 
\midrule[0.7pt]
table& dog& horse& motor& person& potplant& sheep& sofa&  train& tv& book& building& cabinet \\
\midrule[0.7pt]
20.4&56.7&44.5&43.4&56.1&30.3&46.6&26.5&44.5&35.3&17.2&31.0&12.3\\
22.6&58.0&47.2&\textbf{46.6}& \textbf{59.1}&29.8&46.6&\textbf{30.0}&49.5& 41.7& \textbf{18.8}& 35.5& 14.0\\
\textbf{22.5}& \textbf{58.3}& \textbf{47.5}& 44.8& 58.3& \textbf{31.5}& \textbf{47.7}& 29.5& \textbf{49.6}& \textbf{43.5}& 17.5& \textbf{35.6}& \textbf{14.4}\\
\midrule[0.7pt]
ceiling& cup& fence& floor& food& grass& ground& keyboard& light& mountain& mouse& curtain& platform\\
\midrule[0.7pt]
31.5& 11.2& 17.0& 30.6& 27.4& 52.3& 31.1& 31.0& 12.4& 29.0& \textbf{13.9}& 17.6& 20.8\\
31.7& 13.2& 20.2& 34.3& 29.4& 59.1& \textbf{33.3}& 30.2& \textbf{16.2}& \textbf{29.3}& 9.6& 17.4& 17.0\\
\textbf{32.0}& \textbf{13.7}& \textbf{20.2}& \textbf{34.4}& \textbf{30.4}& \textbf{59.7}& 32.3& \textbf{31.3}& 14.5& 28.8& 11.2& \textbf{18.9}& \textbf{20.9}\\
\midrule[0.7pt]
sign& plate& road& rock& sky& snow& bedclothes& track& tree& wall& water& window& 
\textbf{mIoU} \\
\midrule[0.7pt]
10.4& 16.7& 31.3& 20.5& 71.8& 23.3& 13.1& 38.1& 50.4& 36.2& 58.5& 15.7& 27.9\\
\textbf{13.0}& 19.6& 31.4& 22.8& \textbf{74.1}& 26.6& 12.6& 39.4& 52.6& 38.8& 58.8& 18.9& 29.7\\
12.8& \textbf{20.1}& \textbf{33.7}& \textbf{22.8}& 73.9& \textbf{27.1}& \textbf{13.2}& \textbf{41.4}& \textbf{52.6}& \textbf{39.4}& \textbf{59.0}& \textbf{19.4}& \textbf{30.0}\\
\bottomrule[2pt]
\end{tabular}}
\vspace{-3mm}
\caption{Class-wise quantitative comparisons on PASCAL-Context validation dataset.}
\label{tab:class}
\end{table*}

\begin{figure*}
\centering
\subfigure{
\includegraphics[width=\WIDTHFIVE]{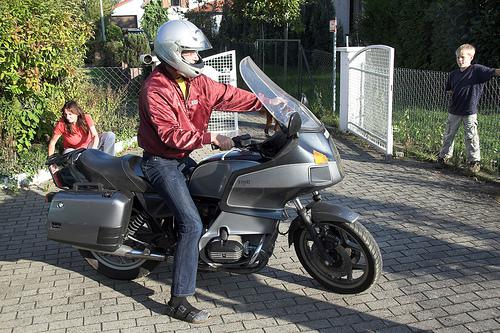}}
\subfigure{
\includegraphics[width=\WIDTHFIVE]{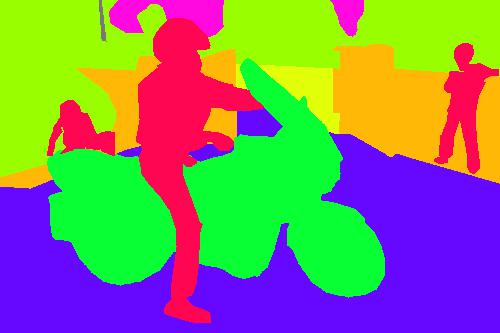}}
\subfigure{
\includegraphics[width=\WIDTHFIVE]{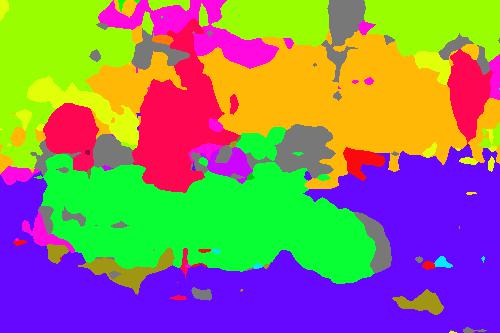}}
\subfigure{
\includegraphics[width=\WIDTHFIVE]{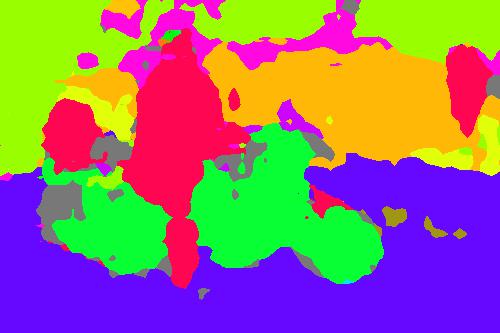}}
\subfigure{
\includegraphics[width=\WIDTHFIVE]{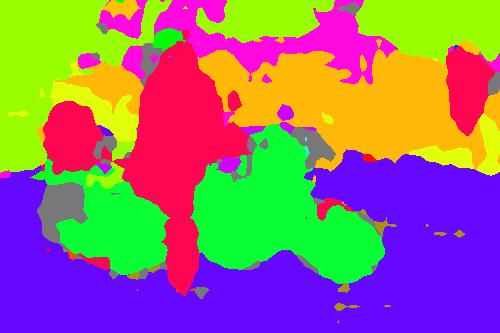}}\vspace{-0.8mm}
\subfigure{
\includegraphics[width=\WIDTHFIVE]{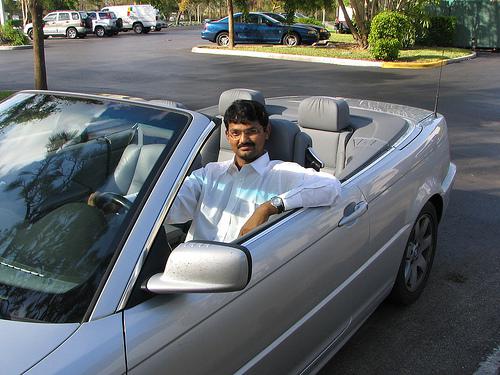}}
\subfigure{
\includegraphics[width=\WIDTHFIVE]{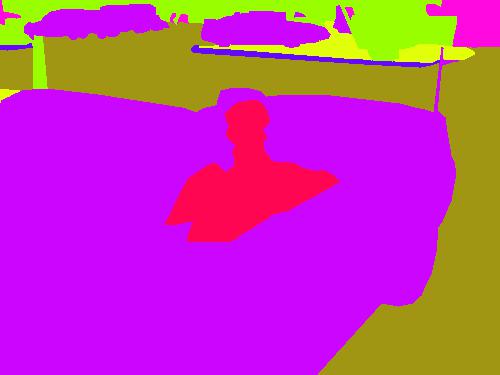}}
\subfigure{
\includegraphics[width=\WIDTHFIVE]{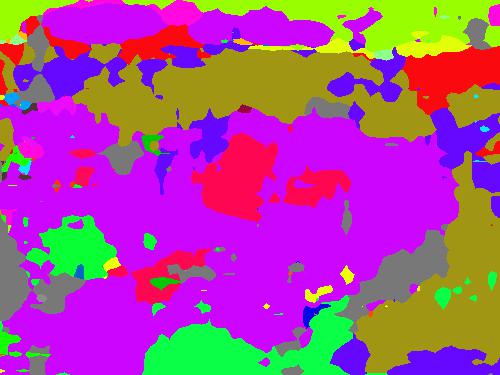}}
\subfigure{
\includegraphics[width=\WIDTHFIVE]{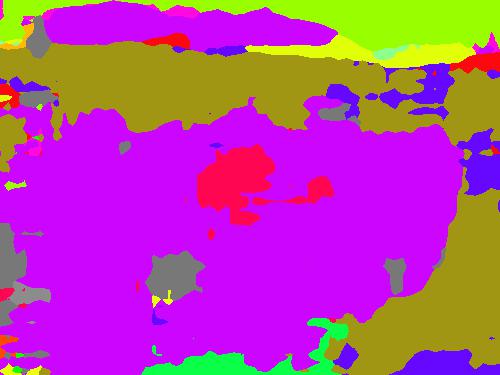}}
\subfigure{
\includegraphics[width=\WIDTHFIVE]{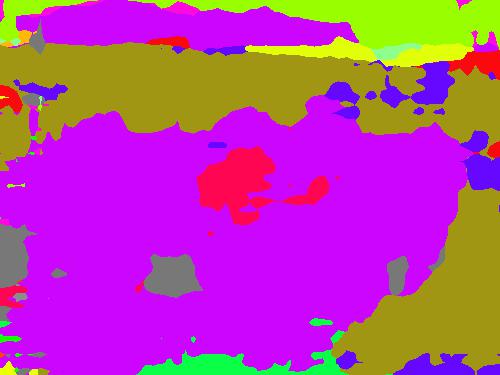}}\vspace{-0.8mm}\addtocounter{subfigure}{-10}
\subfigure[Image]{
\includegraphics[width=\WIDTHFIVE]{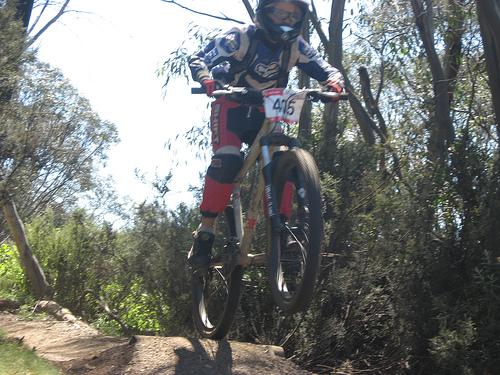}}
\subfigure[Ground Truth]{
\includegraphics[width=\WIDTHFIVE]{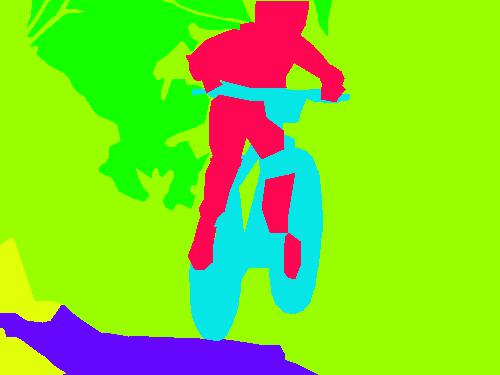}}
\subfigure[PointSup]{
\includegraphics[width=\WIDTHFIVE]{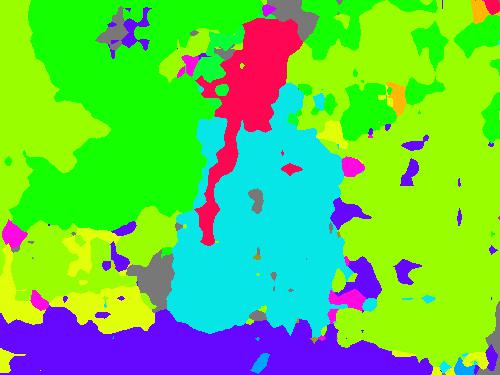}}
\subfigure[PointSup+PDML]{
\includegraphics[width=\WIDTHFIVE]{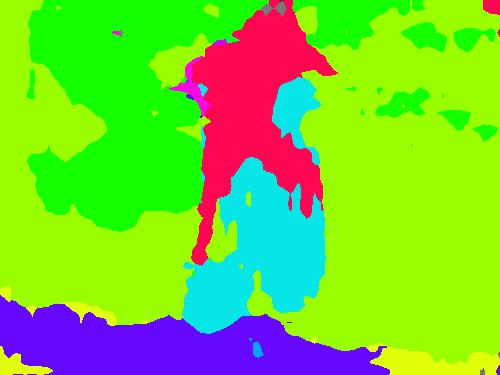}}
\subfigure[Our method]{
\includegraphics[width=\WIDTHFIVE]{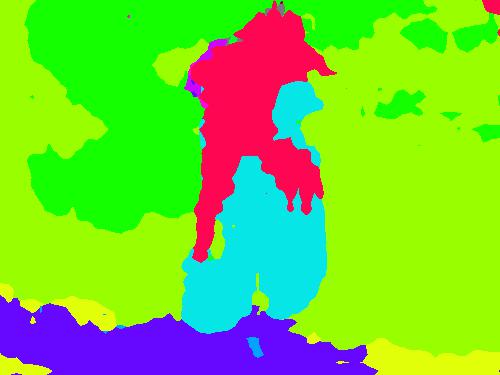}}
\vspace{-3mm}
\caption{Qualitative comparison of different methods on PASCAL-Context validation dataset.}
\label{fig:compare1}
\end{figure*}

\begin{figure*}
\centering
\subfigure{
\includegraphics[width=\WIDTHFIVE]{quality/6_1.jpg}}
\subfigure{
\includegraphics[width=\WIDTHFIVE]{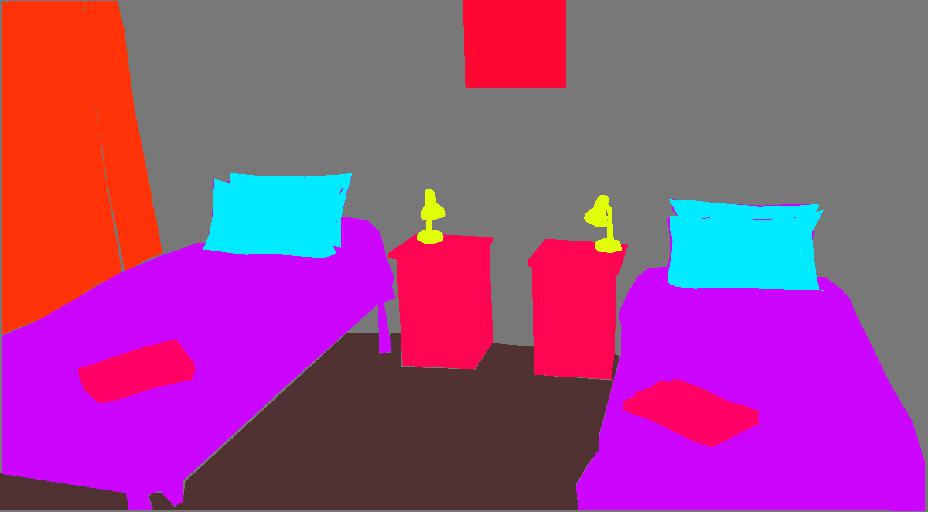}}
\subfigure{
\includegraphics[width=\WIDTHFIVE]{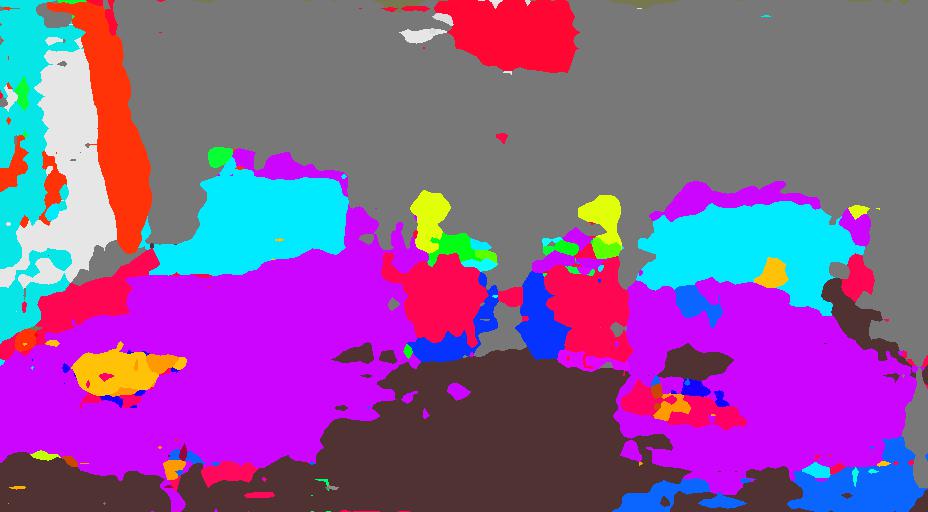}}
\subfigure{
\includegraphics[width=\WIDTHFIVE]{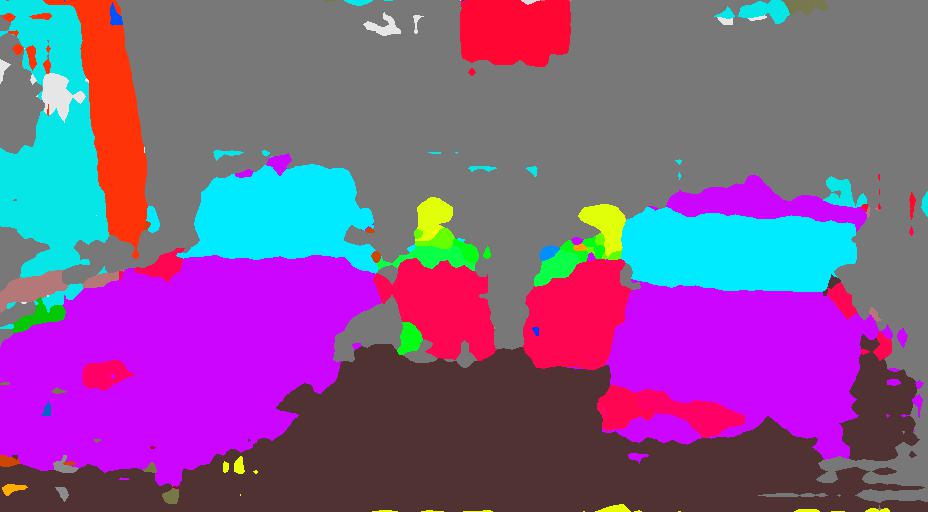}}
\subfigure{
\includegraphics[width=\WIDTHFIVE]{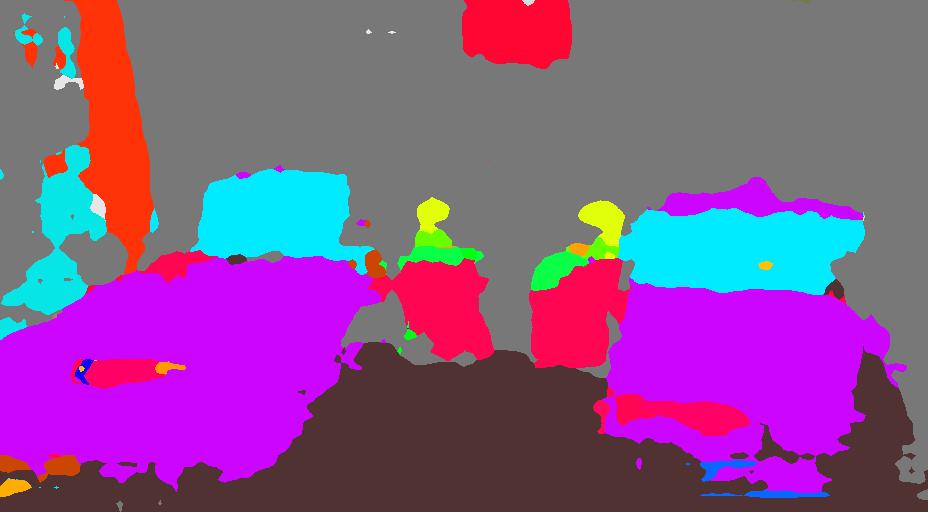}}\vspace{-0.8mm}\addtocounter{subfigure}{-5}
\subfigure[Image]{
\includegraphics[width=\WIDTHFIVE]{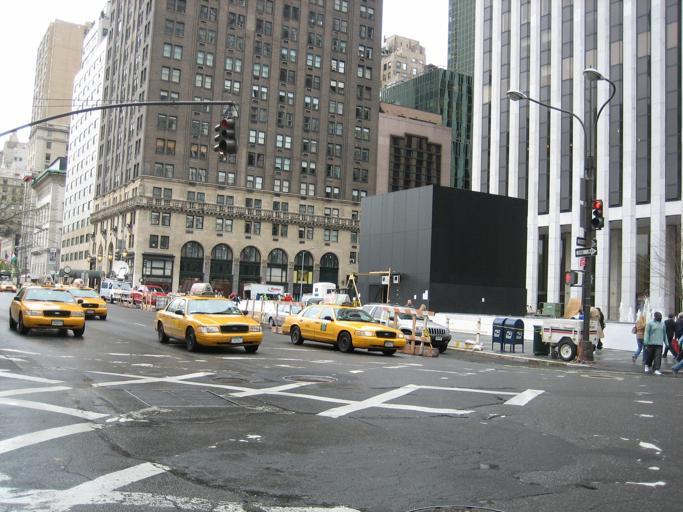}}
\subfigure[Ground Truth]{
\includegraphics[width=\WIDTHFIVE]{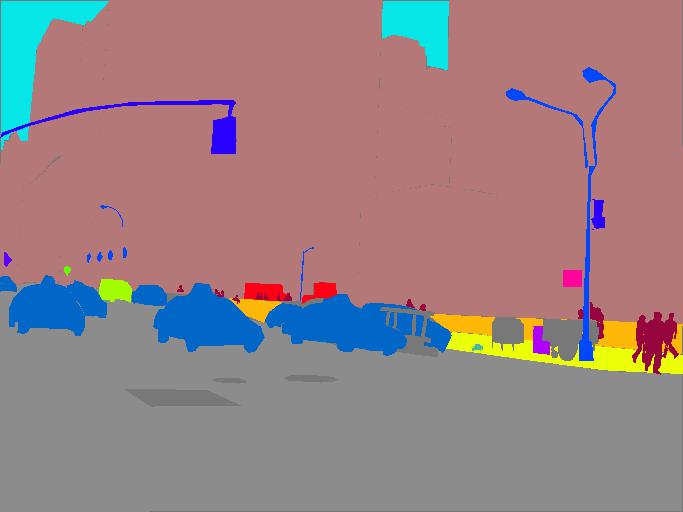}}
\subfigure[PointSup]{
\includegraphics[width=\WIDTHFIVE]{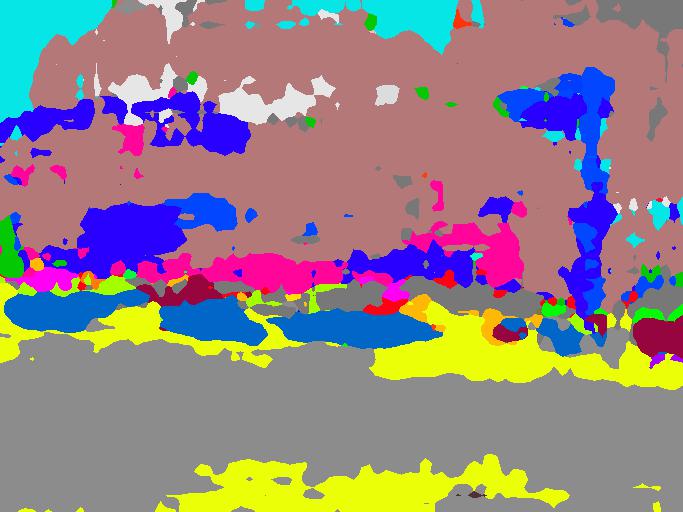}}
\subfigure[PointSup+PDML]{
\includegraphics[width=\WIDTHFIVE]{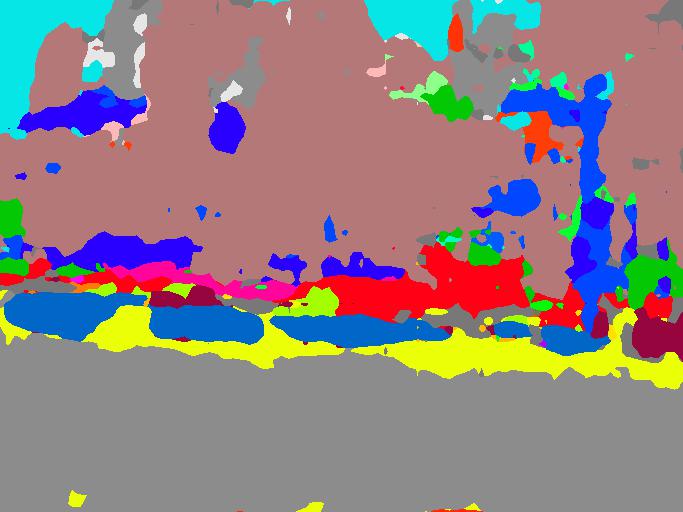}}
\subfigure[Our method]{
\includegraphics[width=\WIDTHFIVE]{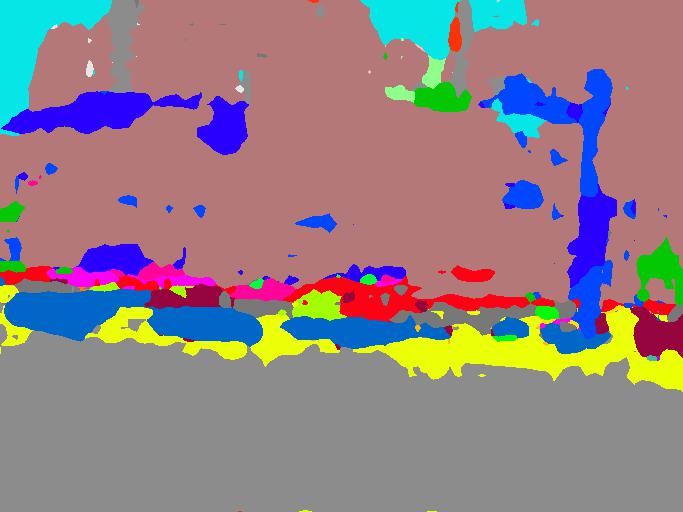}}
\vspace{-3mm}
\caption{Qualitative comparison of different methods on ADE 20K validation dataset.}
\label{fig:compare2}
\end{figure*}

\vspace{-3mm}
\section{Conclusion}
This paper is the first to tackle the task of point-based semantic scene parsing. We propose a novel deep metric learning method to leverage semantic relationship among the annotated points by encouraging the feature representation of the intra- and inter-category points to keep consistent. Points within the same category are optimized to share more similar feature representations and oppositely, those of points from different categories are optimized to be more distinct. Different from all current weakly supervised methods whose solutions are constrained in a single image, our proposed method focuses on optimizing the embedding vectors across different images in the training dataset to obtain sufficient balanced embedding vector pairs. The whole model can be trained in an end-to-end manner. Our method has competitive performance both qualitatively and quantitatively on PASCAL-Context and ADE 20K scene parsing datasets.

\small\noindent \textbf{Acknowledgements.} 
This work is supported in part by IBM-ILLINOIS Center for Cognitive Computing Systems Research (C3SR) - a research collaboration as part of the IBM AI Horizons Network.

\bibliographystyle{aaai}
\bibliography{ref}
\end{document}